\documentclass[10pt, runningheads]{llncs}

\usepackage{graphicx}
\usepackage{booktabs}
\usepackage{textcomp}

\begin{document}
\title{Digital Peter: New Dataset, Competition and Handwriting Recognition Methods}

\author{Mark Potanin\inst{1,2} \and
Denis Dimitrov\inst{1,3} \and
Vladimir Bataev\inst{5,6} \and
Alex Shonenkov\inst{1} \and
Denis Karachev\inst{4} \and
Maxim Novopoltsev\inst{1}\and
Andrey Chertok\inst{1, 7}} 
\authorrunning{M. Potanin et al.}
\institute{Sber, Russian Federation,  
\email{potanin.m.st@sberbank.ru,  dimitrov.d.v@sberbank.ru, shonenkov@phystech.edu, MYNovopoltsev@sberbank.ru}\\
\and
Moscow Institute of Physics and Technology, Russian Federation, 
\email{mark.potanin@phystech.edu}\\
\and
Lomonosov MSU, Russian Federation
\email{denis.dimitrov@math.msu.su}
\and
OCRV, Russian Federation, 
\and 
STC-innovations, Russian Federation,  
\email{bataev@speechpro.com}\\
\and 
University of London, UK
\and
AIRI, Russian Federation,  
\email{Chertok@2a2i.org}\\
}
\maketitle
\begin{abstract}
This paper presents a new dataset of Peter the Great's manuscripts and describes a segmentation procedure that converts initial images of documents into lines. This new dataset may be useful for researchers to train handwriting text recognition models as a benchmark when comparing different models. It consists of 9694 images and text files corresponding to different lines in historical documents. The open machine learning competition "Digital Peter" was held based on the considered dataset. The baseline solution for this competition and advanced methods on handwritten text recognition are described in the article. The full dataset and all codes are publicly available.
\end{abstract}


\section{Introduction}
Text recognition is an area in which deep learning has proven itself perfectly. 
However, the situation in handwritten text recognition (HTR), especially with historical documents, are much worse \cite{gruning2018read,murdock2015icdar}. Since there are only a few open datasets, the quality of trained models' recognition is much lower (see comparison in Table \ref{tab:clovaai-benchmark-result-table-2}).

In this paper, we present a new dataset - Digital Peter - consisting of Peter the Great's manuscripts between 1709 and 1713 and a basic solution for their recognition with deep learning models.

Peter the Great, the first emperor of the Russian Empire who lived at the turn of the 17th and 18th centuries, is one of the most famous figures in Russian history. Peter's handwritten heritage is extensive and diverse and includes his handwritten letters, notes, decrees, and statutes, often with numerous corrections and chroniclers' small additions and marks. 
%
A significant part of the epistolary heritage of Peter the Great is still not introduced to the scientific community. For these reasons, the problem of reading and transcribing his manuscripts continues to remain relevant. 

To estimate the complexity of decoding this dataset, a Digital Peter competition was held, in which more than 200 participants provided their solutions. The best model ran on the Peter, IAM and Bentham datasets. These metrics are demonstrated in the article. We have shown that the Digital Peter dataset can be used as a benchmark for various HTR models. 

The article also analyses the work of models that have proven themselves in optical character recognition (OCR) tasks \cite{R2AM,RARE,GRCNN,STARNet} using both historical and modern handwritten texts. Studies have shown a significant deterioration in quality when it comes to HTR. The remaining paper is organized as follows. Section \ref{related_works} describes the related works and methods of HTR. Section \ref{data_preparing} introduces and describes the details of a new dataset - Digital Peter. Section \ref{section_baselines} includes proposed methods to solve the line segmentation and line recognition tasks and discusses why one may consider the new dataset as a HTR benchmark. Next, Section \ref{advanced_method} describes the competition and suggests some additional techniques that can significantly improve the recognition accuracy. Section \ref{conclusion} concludes the paper.

\section{Related Works}\label{related_works}
\subsection{Datasets}
The Bentham Collection \cite{causer2018making} is a dataset of manuscripts written by Jeremy Bentham - an English philosopher, jurist, and social reformer (1748-1832). 
The IAM Handwriting Database \cite{marti2002iam} contains 1539 pages of handwritten English text from 657 different writers.
In paper \cite{shen2020large}, authors present the HJDataset, a large dataset of historical Japanese documents. It contains more than 2100 pages and over 250,000 layout element annotations. 

As we know, there are no currently free available datasets of Cyrillic historical manuscripts of the 17-18th centuries. There is one dataset \cite{nurseitov2020hkr} of modern Russian and Kazakh language, which consists of more than 1400 forms filled by around 200 different people. And another dataset is the bilingual Evenki/Cyrillic manuscripts by Konstantin Rychkov dated 1911-1913, following pre-reform Cyrillic orthography\footnote{Saint Petersburg Institute of Oriental Manuscripts of RAS, \url{www.orientalstudies.ru}}. It comprises 581 half-pages (59300 words).

\subsection{Methods}

The first HTR models were based on hidden Markov models \cite{hmm2014icacci} and SVM \cite{svm2017spin}. The multi-dimensional LSTM model and CTC-loss \cite{hannun2017sequence} were used in \cite{leifert2014citlab}. The Rosetta \cite{DBLP:journals/corr/abs-1910-05085} uses ResNet as a feature extractor and CTC-loss as a char predictor. Also, for sequence modelling we added recurrent layers in CRNN, \cite{CRNN} and GRCNN \cite{GRCNN} models. It is worth mentioning that the Attention can be used as a char sequence predictor in R2AM \cite{R2AM} and STAR-Net \cite{STARNet}. The last mentioned model contains thin-plate spline (TPS) transformation, which is a variant of the spatial transformation network \cite{NIPS2015_33ceb07b}.

Finding areas of handwritten text is document layout analysis task. These models help to separate the main text from the background. Algorithms such as \cite{6819023} use connected components extraction. 
CNN methods in \cite{renton2018fully} and \cite{Gr_ning_2019} perform pixel-wise classification (between text and non-text labels). A more complex method is to formulate layout analysis as an instance segmentation task \cite{prusty2019indiscapes}.

\section{Data preparation}\label{data_preparing}
\subsection{Data Collection}
The working group was formed at the St. Petersburg Institute of History at the Russian Academy of Sciences, and was represented by historians specializing in Peter the Great's era, paleography and archeography. The data sources for this project were manuscripts from the collections of the St. Petersburg Institute of History at the Russian Academy of Sciences and the Russian State Archive of Ancient Acts (hereinafter - RGADA). The main selection criterion was the presence of a text handwritten by Peter, ranging from two or three words to several pages. FAAR\footnote{Federal Archival Agency of Russia, \url{archives.gov.ru}} and RSAAA\footnote{Russian State Archive of Ancient Acts, \url{rgada.info}} helped and supported the working group by providing digital copies of the manuscripts.

\subsection{Segmentation}
\begin{figure}[ht]
    \centering
    \includegraphics[width=0.5\linewidth]{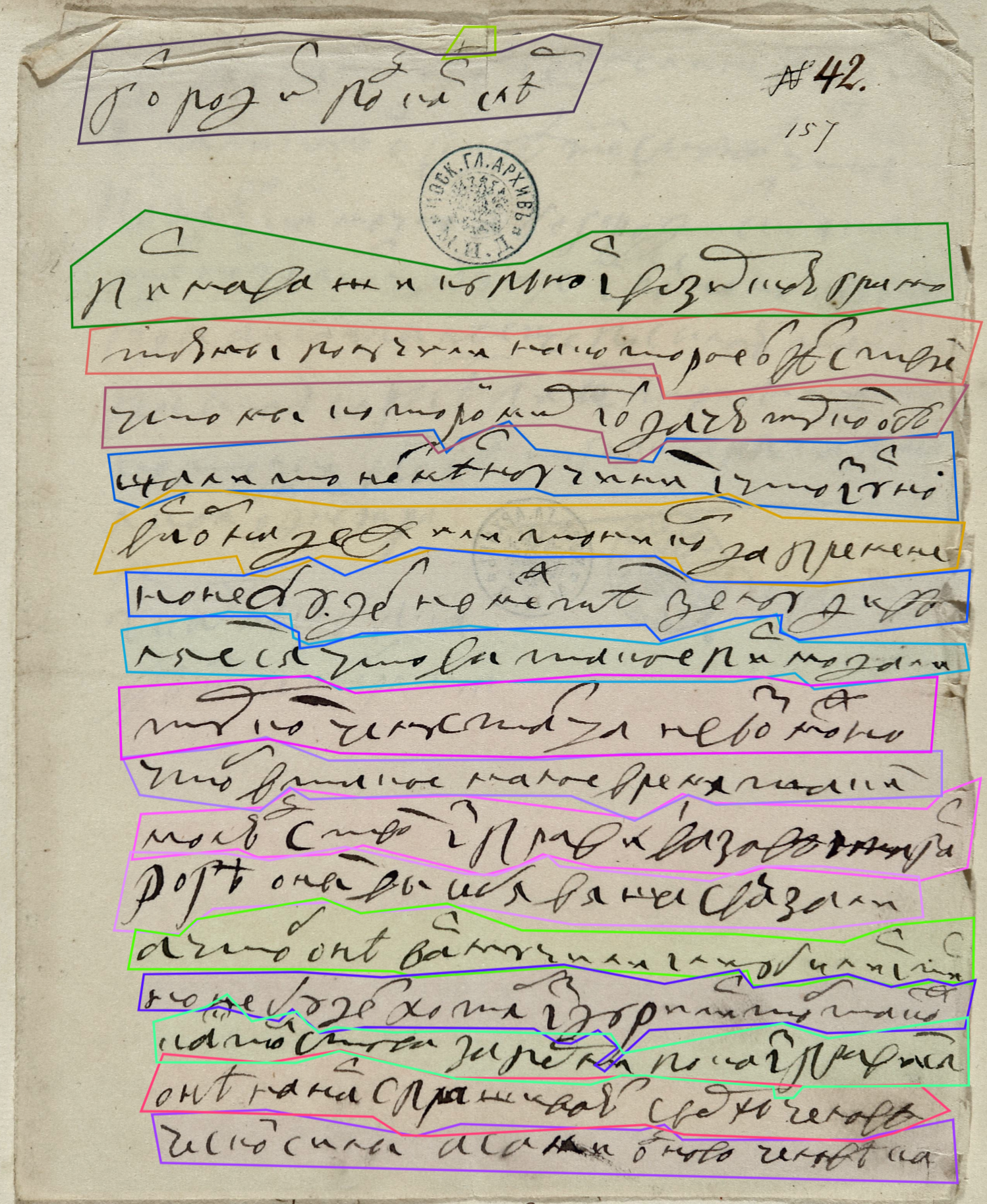}
    \caption{Example of line segmentation marking in CVAT for a single page of the manuscript by Peter the Great.}
    \label{fig:line_label}
\end{figure}

We transcribed Peter's handwritten texts from two data sources:
\begin{enumerate}
    \item Handwritten text images (Fig. \ref{fig:line_label}) of Peter's letters, decrees, and notes in the jpg format. These images had high resolution, approximately $3300\times5400$ ($W \times H$).
    
    \item Text transcribed line by line. Each image from the previous source had a corresponding Microsoft Word file with the text transcription.
\end{enumerate}
Our HTR model took one line of a text from an image as input (Fig. \ref{fig:pipeline}). Therefore, we had to split each page into lines. We did this segmentation task manually, with the Computer Vision Annotation Tool (CVAT) \cite{boris_sekachev_2020_4009388}. In this software assessors do text lines segmentation polygons. But the problem was to map segmented line to corresponding transcription from text document. Therefore, in CVAT we assigned a label for each segmented line corresponding to the line number from the text document (Fig. \ref{fig:line_label}).

\subsection{Lines Cut} \label{cvat}
After segmentation we had the original images, annotations in COCO \cite{DBLP:journals/corr/LinMBHPRDZ14} format, and the transcribed text. Each annotation contained a segmentation mask for a line in polygon format. Also, an annotation contained the information about a bounding box for each mask. These coordinates were automatically received from the edge points of a segmentation mask.

\begin{figure}[ht]
\includegraphics[width=0.8\textwidth]{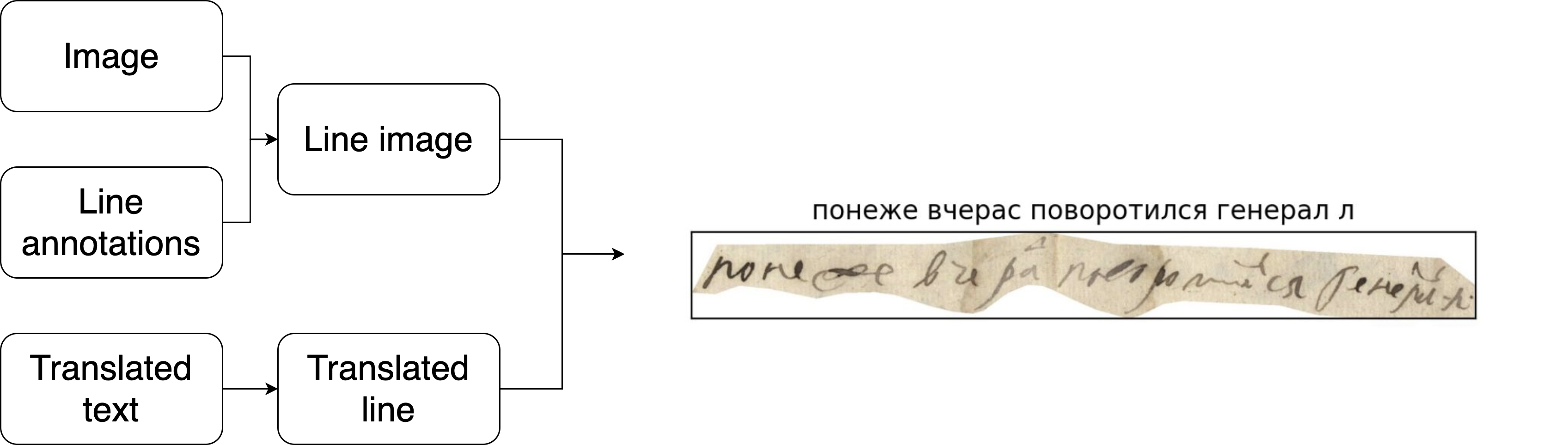}
\caption{Data preparation pipeline.} \label{fig:pipeline}
\end{figure}

To create the final dataset, we had to cut segmented lines from the original image and map them to the corresponding lines in the transcribed text document. The pipeline is presented in Fig. \ref{fig:pipeline}. We cut the lines using the bounding box coordinates from the annotations.



With these coordinates, we cut the rectangle from the original image and put a segmentation mask on it to keep only the segmented line. Pixels outside the mask were set to 255, so we got a white background outside the line we kept. Then we took the line number from the annotations and mapped it to the line in the transcribed text file. Finally we saved the line (rectangle) as a jpg file and the corresponding text as a txt file.



\subsection{Final Dataset}
\begin{figure*}[ht]
 \includegraphics[scale=0.3]{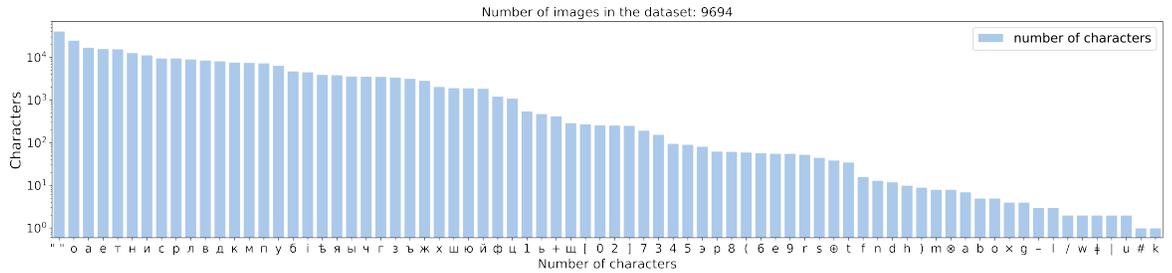}
 \caption{Histogram of characters in the dataset. The first symbol (" ") on horizontal axis is a space symbol.}
\label{fig:counts}
\end{figure*}

Eventually, we had 9694 of image-text pairs. Each pair comprised one image file and one text file. The final dataset is available at \cite{datalink}. There were 265788 characters and approximately 50998 words. The dataset was splitted into three parts: 6237 training pairs, 1930 validation pairs, and 1527 test pairs. A histogram of characters in the dataset is shown in Fig. \ref{fig:counts}. 

\section{Dataset Exploration}\label{section_baselines}

There are two separate tasks which one can try to handle with the described dataset. The first one is \textbf{the line segmentation task}, where the model is trained to segment all the lines in the page. The second one is \textbf{the line recognition task}, where the HTR model training takes place.

\subsection{Line Segmentation Task}
In section \ref{cvat} we mentioned the annotation-creating process in CVAT. Thus, we already had a dataset for the line segmentation model. It took a digitized historical document as input and predicted the masks of lines as the output. With the help of these masks, we then extracted small images of lines from the large image and passed them to our main text recognition model.

We had a total of 662 pages with annotations. Data was split in the following way - 595 images is the train dataset and 67 in the test dataset. The Mask R-CNN R-101-FPN model \cite{he2018mask} was used. It was trained for 9000 iterations. There was only one class that was called "TEXT". This model had a metric mask mAP = 96.7 (IoU=0.5). 

\begin{figure}[ht]
  \centering
  \includegraphics[width=\linewidth]{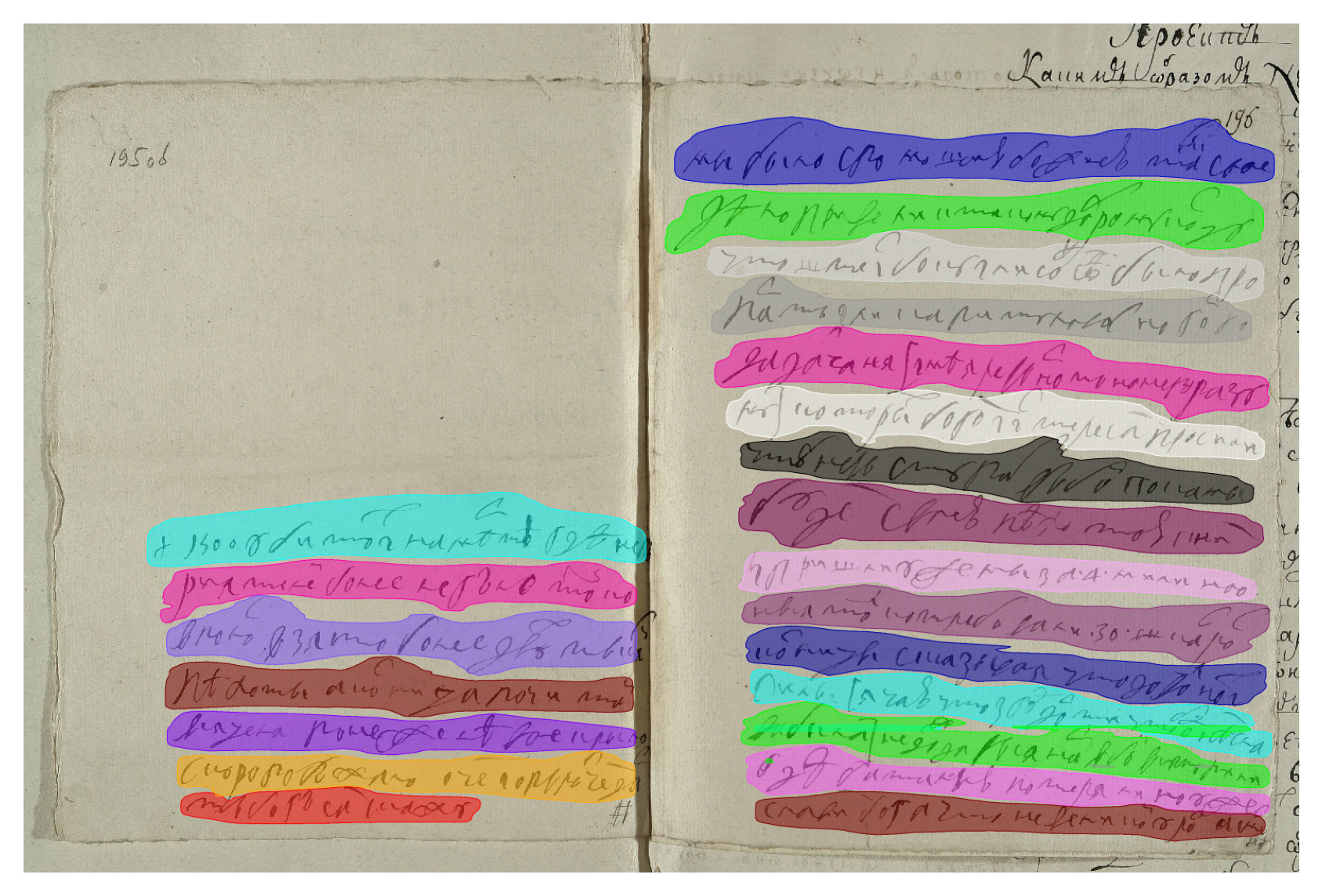}
  \caption{Segmentation model output.}
    \label{fig:segm_model}
\end{figure}

\subsection{Line Recognition Task}

\begin{figure*}[ht]
  \centering
  \includegraphics[width=\linewidth]{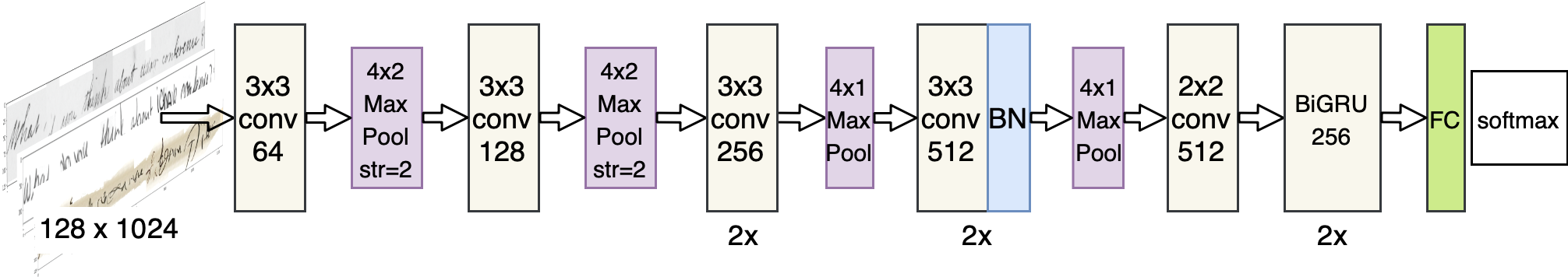}
  \caption{Baseline model architecture.}
    \label{fig:architecture}
\end{figure*}

We came up with the baseline solution for that task. The architecture of the baseline model is represented in Fig. \ref{fig:architecture}. 
All input images had the size $1024 \times 128 \times 3$. After the last $GRU$ layer we had a matrix of shape $255 \times 512$, where $512$ was the feature vector length and $255$ was a sequence of possible characters. After the fully connected layer and softmax activation we had a matrix of shape $255 \times 61$, where $61$ was the number of unique characters. This matrix contained probabilities of possible characters for each position of a string.
After training CRNN, we wanted the model to give us a text on new images; that is, we wanted to get the most probable text from the CRNN matrix. The best path algorithm was used to solve this problem. It consists of the following two steps:
\begin{enumerate}
    \item Calculates the best path by considering the highest probability of a symbol for each position.
    \item Removes spaces and duplicating characters.
\end{enumerate}
Therefore, the baseline model comprised seven CNN layers and two Bidirectional $GRU$ layers and used CTC-loss \cite{CTC}. 

To measure the quality of the HTR models, we used the following metrics: Character Error Rate (\textbf{CER}), Word Error Rate (\textbf{WER}), String Accuracy - exact match of two strings (\textbf{ACC}),and an inference time in seconds on NVIDIA Tesla V100 (\textbf{Time}). 




The metrics of the evaluation of the test part of the dataset is presented in Table \ref{tab:baseline_acc}.

\begin{table}[ht]
  \caption{Baseline metrics}
  \label{tab:baseline_acc}
  \centering
    \begin{tabular}{ |l|c|c|c| }
    \toprule
    \textbf{Model} & \textbf{CER} & \textbf{WER} & \textbf{ACC} \\
    \midrule
    Baseline & 10.5 & 44.4 & 21.7 \\
    \bottomrule
    \end{tabular}
\end{table}

\subsection{The Dataset as a Benchmark}

Handwritten text has several features – letters can have a different appearance depending on the author, the position in the word, the quality of the paper, and significant difference in writing, even for one person. Additionally, historical documents may contain various defects like ink drops or paper defects. Another problem with them is the small amount of data that is usually provided.

\begin{figure}[ht]
 \centering
 \includegraphics[width=\linewidth]{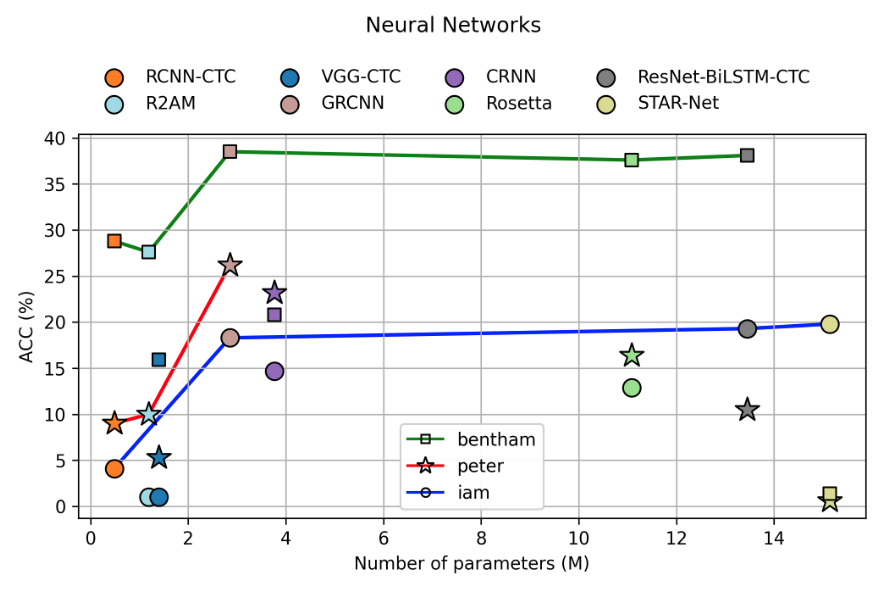}
 \caption{String Accuracy versus number of parameters for some models, the curves demonstrate the most effective models for the selected datasets.}
    \label{fig:clovaai}
\end{figure}

We ran Clova AI models on three handwritten text datasets: IAM, Bentham and Peter. All these models used AdamW optimizer ("weight decay" = $0.01$) with OneCycleLR scheduler, starting at a learning rate of $0.1$ down to {1e-5} with "anneal strategy" = "cos". All pictures were $1024$ pixels in width and $128$ pixels in height. All models were trained for $100$ epochs. A Github repository with all experiments based on the mentioned handwritten datasets is available \cite{Digital-Peter-Model-Comparisons}. The results are represented in Table \ref{tab:clovaai-benchmark-result-table-2} and Fig. \ref{fig:clovaai}.

\begin{table*}[ht]
\begin{center}
\caption{\label{tab:clovaai-benchmark-result-table-2} The best results of training different handwritten recognition models from Clova AI deep text recognition benchmark \cite{clova-dataset}. IAM images have no difficult background unlike historical manuscripts, so we suppose models with attention mechanism require more iterations for Peter and Bentham datasets.}
\begin{tabular}{|l|c|c|c||c|c|c||c|c|c|}
\toprule
\textbf{Model} & \multicolumn{9}{c|}{\textbf{Datasets}} \\
\cmidrule{2-10}
& \multicolumn{3}{c||}{\textbf{IAM}} & \multicolumn{3}{c||}{\textbf{Peter}} & \multicolumn{3}{c|}{\textbf{Bentham}} \\
\cmidrule{2-10}
& \textbf{CER} & \textbf{WER} & \textbf{ACC} & \textbf{CER} & \textbf{WER} & \textbf{ACC} & \textbf{CER} & \textbf{WER} & \textbf{ACC} \\
\midrule
VGG-CTC & 18.4 & 53.1 & 1.0 & 19.3 & 83.1 & 5.3 & 8.5 & 30.2 & 15.9 \\
RCNN-CTC & 12.8 & 39.8 & 4.1 & 14.2 & 72.1 & 9.0 & 6.2 & 22.5 & 28.8 \\
Rosetta \cite{DBLP:journals/corr/abs-1910-05085} & 8.4 & 28.0 & 12.9 & 9.6 & 56.1 & 16.4 & 4.4 & 17.5 & 37.6 \\
CRNN \cite{CRNN} & 7.9 & 24.9 & 14.7 & 8.8 & 44.4 & 23.2 & 8.2 & 26.9 & 20.8 \\
GRCNN \cite{GRCNN} & 7.3 & 22.8 & 18.3 & \textbf{7.7} & \textbf{40.5} & \textbf{26.2} & \textbf{4.2} & \textbf{16.1} & \textbf{38.5} \\
ResNet-BiLSTM-CTC & 6.9 & 21.8 & 19.3 & 15.6 & 64.2 & 10.5 & 4.5 & 17.1 & 38.1 \\
R2AM \cite{R2AM} & 50.3 & 73.0 & 1.0 & 42.2 & 80.6 & 10.0 & 22.1 & 35.5 & 27.6 \\
STAR-Net \cite{STARNet} & \textbf{6.6} & \textbf{21.1} & \textbf{19.8} & 83.8 & 98.5 & 0.6 & 74.7 & 98.8 & 1.4 \\

\bottomrule
\end{tabular}
\end{center}
\end{table*}

\section{Competition}\label{advanced_method}
A large-scale data science competition was held based on the described dataset. The task was prepared in collaboration with the St. Petersburg Institute of History at the Russian Academy of Sciences, FAAR and RSAAA. Nearly $1 000$ teams took part in the competition. Participants were invited to create an algorithm for line-by-line recognition of manuscripts written by Peter the Great. 
The leaderboard considered the following recognition quality metrics: \textbf{CER}, \textbf{WER}, and \textbf{ACC}. \textbf{CER} was a key metric used to sort the leaderboard. The private and public leaderboards and more detailed information about the competition can be found in \cite{complink}. 

We analysed the best solutions provided by participants. Below, we show a few of their methods, that improved the recognition accuracy. 

\subsection{Augmentations, Architecture Improvements, and Optimization Process.}

For the first time, instead of using images of a fixed size, we maintained the aspect ratio of images by using a fixed height of 128 and a variable width. 

Also, since the dataset was quiet small, we added augmentations to improve model robustness. We used the Albumentations \cite{Albumentations} package, and found the following modifications useful:
\begin{itemize}
	\item Random image rotation by $4^{\circ}$ clockwise or counterclockwise'
	\item Random change of width from $95\%$ to $105\%$'
	\item Grid distortion.
\end{itemize}
We made several modifications to the model architecture. We added the Batch-Normalization 
before all the layers and after all the activations in the convolutional part of the model. The model had a left context of 19 and a right context of 23, so we added the corresponding number of vectors to all the images by the $x$ axis using reflection padding. Also, we made the model have a subsampling factor of 4, so that given input width $w$, the output had a width of $w / 4$. We varied the GRU cell size and found out that a GRU with 368 hidden units performs slightly better than one with 256. 

We also discovered that SGD with momentum performs significantly better than Adam
, as shown in Table \ref{tab:Tuning_Models}. The best model with Adam optimizer was trained with an exponential learning rate decay from $10^{-3}$ to $10^{-5}$ for 32 epochs. 

The training model with the SGD optimizer was a bit trickier. Since CTC-loss is numerically unstable, especially during first epochs, starting with a high learning rate is impossible, and starting with a low learning rate led to poor model quality. In such cases, the cyclical learning rate schedule \cite{smith2017cyclical} turned out to be useful: for the first four epochs we linearly increased learning rate from $10^{-6}$ to $10^{-2}$, and then decreased it to $10^{-6}$ again. Using this method with tha AdamW optimizer showed poor results. Nevertheless, it significantly improved the models, optimized with SGD. All models were trained using a weight decay of $10^{-2}$.

\begin{table}[ht]
  \caption{Tuning Models}
  \label{tab:Tuning_Models}
  \centering
    \begin{tabular}{ |l|c|c|c|c|c| }
    \toprule
    \textbf{Model} & \textbf{GRU cell} & \textbf{Optimizer} & \textbf{CER} & \textbf{WER} & \textbf{ACC} \\
    \midrule
    CRNN & 256 & Adam & 7.1 & 39.7 & 24.9 \\
    CRNN & 368 & Adam & 6.6 & 37.2 & 29.0 \\
    CRNN & 368 & SGD & \textbf{5.7} & \textbf{33.1} & \textbf{33.1} \\
    \bottomrule
    \end{tabular}
\end{table}

\subsection{Beam Search Decoding with Language Model}

For decoding we used Beam Search \cite{CTCDecoder} with statistical n-gram language model (LM). A similar pipeline with word-level LMs is widely used for speech recognition with CTC-loss\cite{DeepSpeech2014}. 
In this task, simple closed-vocabulary decoding seemed to be useless: vocabulary, based on $93\%$ of randomly selected training data, indicated more than $20\%$ out-of-vocabulary (OOV) rate on other $7\%$ of texts. Also, we had to minimize the CER. For that task, a word-level LM was suboptimal, since decoding with a word-level model optimizes WER.
Hence, we used a 6-gram character-level model with all possible characters (including spaces as separating characters). 

All models were built with the SRILM\cite{SRILM} toolkit. We separated 7\% of the data for fine-tuning and selecting models and used the remaining 93\% of data to construct the \textbf{Small} model. 

The decoding itself finds a sequence of characters $c$c, that maximizes the following objective \cite{DeepSpeech2014}:

$$Q(c) = \log{P(c|x)} +\alpha\log{P_{lm}(c)} +\beta symbol\_count(c)$$

where $\alpha$ is an LM weight and $\beta$ is a symbol insertion penalty. 

In most of our experiments we discovered that the model worked perfectly with $\alpha = 0.8$, $\beta = 2.0$, and handling more than 100 active hypotheses does not improve results, but slows down the process. The impact of decoding using the LM is shown in Table \ref{tab:Decoding_impact}.

\begin{table}[ht]
  \caption{Decoding with the language model}
  \label{tab:Decoding_impact}
  \centering
    \begin{tabular}{ |l|c|c|c|c|c| }
    \toprule
    \textbf{Model} & \textbf{Optimizer} & \textbf{LM} & \textbf{CER} & \textbf{WER} & \textbf{ACC} \\
    \midrule
    CRNN, 368 & Adam & - & 6.6 & 37.2 & 29.0 \\
    CRNN, 368 & Adam & Small & \textbf{4.2} & \textbf{22.6} & \textbf{47.3} \\
    \midrule
    CRNN, 368 & SGD & - & 5.7 & 33.1 & 33.1 \\
    CRNN, 368 & SGD & Small & \textbf{3.7} & \textbf{20.7} & \textbf{50.1} \\
    \bottomrule
    \end{tabular}
\end{table}

To construct the final \textbf{Large} LM, we interpolated the \textbf{Small} model, based on training texts, with the model and constructed historical texts of the 17th century from GramEval 2020 \cite{grameval2020} corpus. Decoding with the \textbf{Large} model further improves CER, as shown in Table \ref{tab:Final_Models}.

\begin{table}[ht]
  \caption{Final Models}
  \label{tab:Final_Models}
  \centering
    \begin{tabular}{ |l|c|c|c|c|c| }
    \toprule
    \textbf{Model} & \textbf{Optimizer} & \textbf{LM} & \textbf{CER} & \textbf{WER} & \textbf{ACC} \\
    \midrule
    CRNN, 368 & SGD & - & 5.7 & 33.1 & 33.1 \\
    CRNN, 368 & SGD & Small & 3.7 & 20.7 & 50.1 \\
    CRNN, 368 & SGD & Large & \textbf{3.5} & \textbf{19.4} & \textbf{52.3} \\
    \bottomrule
    \end{tabular}
\end{table}

\section{Conclusion}\label{conclusion}

In this paper, we presented the handwritten dataset of Peter the Great's manuscripts. This is expected to be useful for research in HTR. The dataset consists of the text lines written by Peter's hand. Almost all the words are outdated and are no longer used in the modern Russian. An open machine learning competition "Digital Peter" was based on the above mentioned dataset. Newly developed models showed impressive quality on the text recognition task. Also, using new values of hyperparameters for GRU layers and optimizer and using Beam Search with large n-gram language model for decoding outputs of CRNN  we got better values of CER for the presented dataset than our attempts with Clova AI deep text recognition benchmark.

As we elaborate our work in the future, we plan to collect more archival manuscripts written by other historical figures as well. We also plan to eventually develop a software that can transcribe digitized archival documents. 

\bibliographystyle{splncs04}
\bibliography{camera-ready-scheme}

\end{document}